\pdfoutput=1

\documentclass[11pt]{article}

\usepackage[preprint]{acl}

\usepackage{times}
\usepackage{latexsym}
\usepackage{longtable}

\usepackage[T1]{fontenc}

\usepackage[utf8]{inputenc}

\usepackage{microtype}

\usepackage{inconsolata}

\usepackage{graphicx}

\usepackage{array}
\usepackage{inconsolata}
\usepackage{multirow}
\usepackage{subcaption}
\title{Mission Impossible: Feedback-Guided Dynamic Interactive Planning for Improving Reasoning on LLMs}

\author{
    Dong Yan\textsuperscript{1}, Gaochen Wu\textsuperscript{2}, Bowen Zhou\textsuperscript{2,3}\thanks{Corresponding Author}\\
    \textsuperscript{1}Central South University\\
    \textsuperscript{2}Tsinghua University\\
    \textsuperscript{3}Shanghai AI Laboratory\\
    \texttt{yandong203519@gmail.com}
}

\begin{document}
\maketitle
\begin{abstract}
Recent advancements in language agents have led to significant improvements in multi-hop reasoning tasks. However, existing approaches often struggle with handling open-domain problems, which require massive information retrieval due to their reliance on a fixed sequence of actions. To address this, we propose Feedback-Guided Dynamic Interactive Planning (FGDIP), a novel framework tailored to enhance reasoning in LLMs by utilizing dynamic and adaptive strategies for information exploration in open-domain multi-hop reasoning tasks. Our approach begins by identifying key entities relevant to the problem, which serve as the initial nodes in the reasoning process. From these initial nodes, we then generate reasoning child nodes with the process being refined through a combination of historical error analysis and real-time feedback, which allows the framework to dynamically adjust and optimize its reasoning strategies. By integrating depth-first search with an innovative node generation technique, our framework adapts based on both prior error paths and concurrently generated nodes at the same hierarchical level. This dynamic strategy effectively expands the search space while ensuring the reasoning process systematically converges toward accurate solutions.
Experimental results show that FGDIP achieved up to 54.47\% F1 score on the HotpotQA dataset and 70.05\% on the StrategyQA dataset, surpassing the best baseline by 5.03\% and 7.25\% respectively, highlighting its versatility and potential to enhance language agents in multi-hop reasoning tasks.
\end{abstract}

\begin{figure*}[t]
    \centering
    \includegraphics[width=1.0\textwidth]{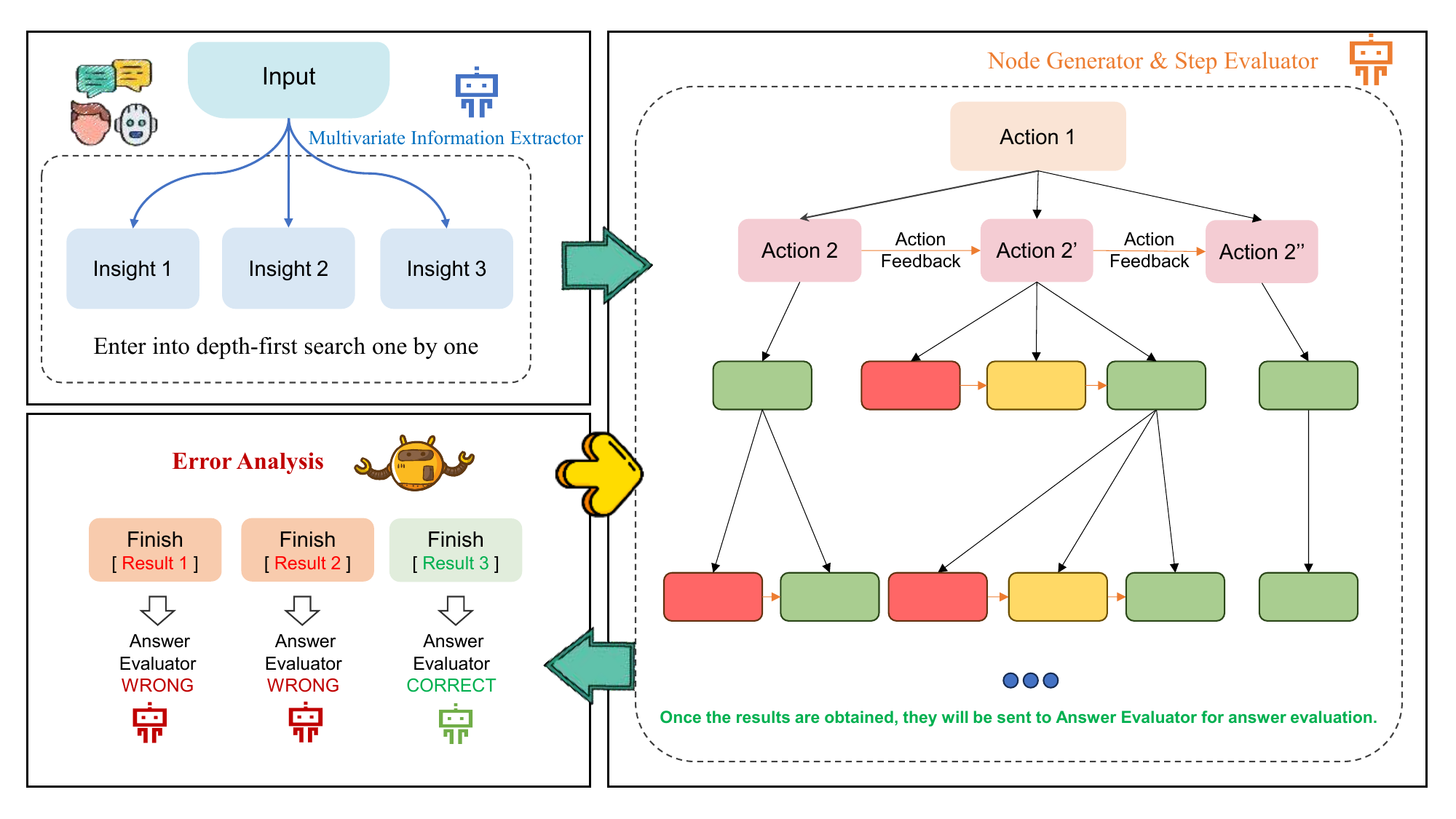}  
    \caption{Overview of the FGDIP framework. It initiates by identifying key entities related to the problem and utilizes a depth-first search approach for thorough exploration of each node. The process draws insights from both error antecedent branches and concurrently generated nodes at the same hierarchical level, thus facilitating the refinement and optimization of subsequent exploration.The Step Evaluator assesses the feasibility of each newly generated node, provides a detailed rationale, and then ranks nodes to prioritize the most promising paths. Answer Evaluator assesses the results for relevance, integrating any error analysis to enhance future exploration.}
    \label{fig:label}
\end{figure*}

\section{Introduction}

The emergence of language agents marks a significant milestone in the evolution of artificial intelligence technologies \citep{Wang_2024,xi2023rise}. Built on robust LLM frameworks like GPT-4 \citep{openai2024gpt4}, these agents handle complex tasks by generating real-time strategies. Recent research emphasizes improvements in task planning and execution \citep{wei2023chainofthought,yao2022react} through stepwise problem-solving. Studies have explored adaptive learning techniques, which are crucial for agents to reflect on past interactions, learn from experiences, and improve future performance \citep{shinn2023reflexion,madaan2023selfrefine}. Moreover, collaboration among multiple agents has been enhanced, allowing for the division of tasks among specialized systems \citep{hong2023metagpt,chen2023agentverse}. Integrating external tools has expanded language agents' scope, enabling access to diverse information and computational resources for nuanced responses \citep{yao2022react,lu2023chameleon,qin2023toolllm}.

Although these methods demonstrate notable effectiveness, they remain challenged in open-domain tasks such as HotpotQA \citep{yang2018hotpotqa}, where dynamic information retrieval and data filtering play a pivotal role in supporting multi-hop question answering. In addressing multi-hop reasoning challenges, prevalent methodologies \citep{zhu2024knowagent,qiao2024autoact,han2024uncertaintyawarelanguageagent,gao2024dr3} that utilize ReAct \citep{yao2022react} paradigm for sequential execution demonstrate inherent limitations. Initially, these models perform a breakdown of the problem, establishing a retrieval sequence that rigidly guides subsequent searches. This linear, token-by-token decision-making process, once initiated, constrains the model's ability to dynamically adapt its strategy based on evolving information. When essential contexts are missing during search iterations, models are more likely to deem a solution unattainable or repeatedly search the same keywords until they exhaust all attempts. These limitations constrain the search space of these methods when solving open domain problems. Additionally, current approaches in natural language processing often involve fine-tuning language models \citep{zhu2024knowagent,qiao2024autoact}, which is computationally expensive and data-intensive.
\\ 
\hspace*{1em}Motivated by the need for effective multi-hop reasoning in open-domain problems, we propose FGDIP, a novel framework that dynamically adapts its strategies for information exploration without requiring any model fine-tuning. As shown in Figure~\ref{fig:label}, the model begins by identifying key entities relevant to the problem, which serve as the initial nodes in the reasoning process. Each node represents distinct thought, action, and observation step. The model employs a depth-first search strategy to explore each initial node thoroughly. During this process, newly expanded nodes incorporate insights from both error antecedent branches and concurrently generated nodes at the same hierarchical level, which ensures comprehensive exploration of the solution space and continuous refinement of strategies towards converging on the correct answer. The Step Evaluator assesses the feasibility of each newly generated node, provides a detailed rationale, and then ranks the node among others to prioritize the most promising paths. Finally, upon deriving a result, Answer Evaluator examines the result to determine its relevance to the topic. If the result is deemed on-topic, the process concludes; otherwise the experience from erroneous judgments is integrated for future reference.

Our experimental evaluation reveals that FGDIP significantly surpasses existing baselines on both the HotpotQA \citep{yang2018hotpotqa} and StrategyQA \citep{10.1162/tacl_a_00370} datasets. On HotpotQA, FGDIP achieves 60.46\% F1 score on easy questions, 53.87\% on medium questions, and 48.56\% on hard questions, representing improvements of 3.08\%, 1.80\%, and 8.66\% respectively over the best baseline. On StrategyQA, FGDIP achieves an F1 score of 70.05\%, outperforming UALA \citep{han2024uncertaintyawarelanguageagent} by 7.25\%. Additionally, FGDIP also demonstrates strong performance on the Game of 24 dataset, a mathematical reasoning challenge regarded as a closed-domain task. These results demonstrate FGDIP's superior ability to dynamically adjust its reasoning strategies and effectively handle complex, multi-hop, and implicit reasoning tasks. The main contributions are as follows:

\begin{itemize}
    \item We propose FGDIP, a novel framework for solving open-domain problems that require multi-hop reasoning, which expands the scope of information search and employs more flexible and adaptive strategies.
    \item  We propose a method in which each step of the decision-making process leverages the accumulation of historical error analysis and a comprehensive real-time action feedback. This approach enables the model to initially diverge and subsequently converge on the correct answer throughout the multi-hop reasoning process.
    \item We conduct comprehensive experiments of our framework, demonstrating that FGDIP can improve the downstream performance of multi-hop reasoning tasks in both open and closed domain tasks.
\end{itemize}

\section{Related Work}
\subsection{Language Agents}

As the field of language agents \citep{Wang_2024} advances, various strategies have emerged to leverage their reasoning capabilities \citep{qiao2023reasoning}. ReAct \citep{yao2022react} combines reasoning and action to tackle complex tasks, adapting to real-time interactions. Reflexion \citep{shinn2023reflexion} enhances language agents through verbal feedback, avoiding traditional weight updates. Chameleon \citep{lu2023chameleon} offers a flexible framework that equips LLMs with diverse tools for compositional reasoning across a wide range of problems. Dr3 \citep{gao2024dr3} offers a systematic approach to detecting and rectifying off-topic answers through dynamic reasoning refinement in multi-hop question answering. The KnowAgent \citep{zhu2024knowagent} framework enhances task planning and execution using a knowledge-augmented approach to minimize planning hallucinations and improve action coherence. UALA \citep{han2024uncertaintyawarelanguageagent} introduce a language agent framework that leverages uncertainty measurement to refine reasoning processes and optimize external resource utilization. Unlike their work, our approach involves more comprehensive and adaptive strategies designed to enhance the exploration of information, enabling the model to dynamically adjust its strategies based on available context, rather than being constrained to a fixed sequence of actions.

\subsection{Multi-Step Reasoning}

Recent advancements in large language models have highlighted the efficacy of structured prompting techniques for enhanced reasoning. The Chain-of-Thought (CoT) \citep{wei2023chainofthought} approach enables LLMs to tackle complex problems by sequentially following thought processes. The Self-Consistency \citep{wang2023selfconsistency} strategy builds on this by generating multiple reasoning chains and selecting the most consistent one. The Tree of Thoughts model \citep{yao2023tree} further advances the chain-of-thought approach by allowing LLMs to explore multiple reasoning paths within a structured framework. Given its high capabilities, the foundational architecture of our approach extensively employs the ToT structure. Nevertheless, our architecture transcends the conventional ToT framework. In the generation of new nodes, our approach systematically incorporates insights from both antecedent branches and concurrently generated nodes at the same hierarchical level, thereby facilitating graph-like interconnections among the nodes. Additionally, the Graph of Thoughts \citep{Besta_2024} framework structures thoughts in a graph format to support intricate and interdependent reasoning, fostering a robust mechanism for solving more elaborate problems. The majority of existing studies focus on multi-hop reasoning within well-defined, closed problem spaces. Conversely, our research is directed towards addressing open-ended problems that necessitate extensive exploration and creative thinking and our methodological approach systematically narrows down potential solutions through a progressive refinement during the exploration process. 
\begin{figure*}[t]
 \vspace{-1.5\baselineskip}
    \centering
    \includegraphics[width=1.0\textwidth]{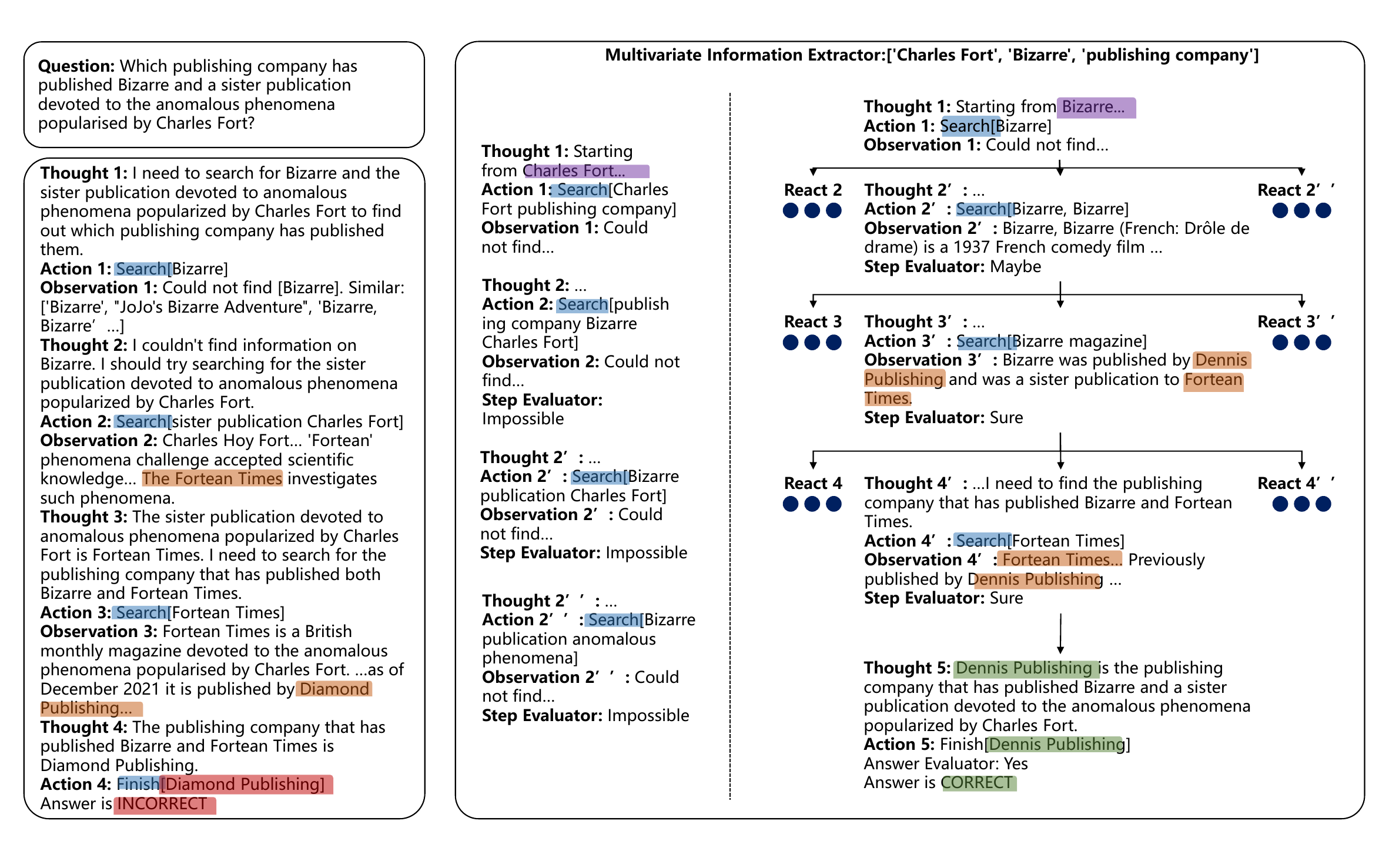}  
    \caption{Comparison between ReAct paradigm and FGDIP. The ReAct method failed due to vague keyword use like "Charles Fort publishing company" and didn't adapt when initial searches were unsuccessful. It also incorrectly identified only a temporary publisher for "Fortean Times" without confirming the publisher for "Bizarre." In contrast, FGDIP efficiently switched to "Bizarre magazine" after initial searches failed, quickly identifying Dennis Publishing as the common publisher for both "Bizarre" and "Fortean Times," demonstrating its precision in query processing.}
    \label{fig:mylabel}
\end{figure*}

\section{Method}
\subsection{Multivariate Information Extractor}
The Multivariate Information Extractor module is designed to identify and extract key entities that are critical for solving complex problems. It starts by analyzing the problem statement to determine which entities are specific and unique, ensuring that the problem is thoroughly understood from multiple dimensions. These extracted entities, serving as the first layer of nodes, facilitate a more targeted and effective strategy for further reasoning.

Traditional models addressing complex queries with multiple entities often follow a rigid sequence of actions, focusing initially on the most obvious entity and attempting to gather all necessary information linearly. This approach can fail if the entity lacks all the relevant data needed for resolution.

In contrast, our method avoids such pitfalls by not committing to a fixed procedural pathway from the outset. After the first attempt proves insufficient for solving the query, our model shifts to exploring alternative entities. By doing so, it opens up new avenues for inquiry that are more likely to lead to successful problem resolution. This iterative, entity-focused strategy enhances the model's ability to handle complex, multi-faceted queries more effectively, ensuring a higher rate of successful information retrieval and problem-solving.

\subsection{Node Generator}
Our method employs adaptive candidate generation, leveraging historical error analysis and real-time feedback to optimize multi-hop reasoning.
\subsubsection{Error Analysis}
To enhance the decision-making process and improve accuracy, our method meticulously maintains a record of past errors resulting from Answer Evaluator, which will be introduced in Section~\ref{subsubsec:answer_evaluator}. The insights derived from these errors, denoted as $e_{1\dots i-1}$, are leveraged to guide and refine future decision-making steps. This historical analysis ensures that the model learns from previous mistakes, thereby preventing the repetition of ineffective strategies and enhancing the overall decision accuracy.
\subsubsection{Real-time Feedback}
In addition to historical analysis, real-time feedback plays a crucial role during node expansion. As the model explores new nodes $z_{ij}$, it continuously evaluates the real-time feedback from the interactions with concurrently generated nodes $z_{ij}^{'}$ at the current layer. This integration is essential for maximizing the utility of the most recent observations, ensuring that useful information is effectively utilized and redundant actions are minimized. The combination of leveraging past errors and real-time feedback ensures a more adaptive and accurate problem-solving approach, leading to a higher success rate in complex reasoning tasks.
\subsubsection{Node Generation}
Combining the insights from historical error analysis and real-time feedback integration, our method formalizes the generation of new nodes in the decision-making process. Specifically, when generating the j-th layer node $z_{ij}$  in the i-th branch, the node $z_{ij}$ is influenced by:
\begin{itemize}
    \item The input question $x$.
    \item The historical error analysis $e_{1\ldots i-1}$ from the preceding branches.
    \item Real-time feedback from interactions with other nodes $z_{ij}^{'}$ at the current layer.
    \item The current branch’s trajectory $z_{i1, \ldots, i(j-1)}$.
\end{itemize}
The node generation approach is formally expressed as:
\[
z_{ij} \sim p(z_{ij} \mid x, e_{1\ldots i-1}, z_{ij}^{'}, z_{i1, \ldots, i(j-1)}).
\]

\subsection{Evaluator}
Inspired by \citet{gao2024dr3}, our method enhances the evaluation process by introducing a detailed and explanatory framework for assessing both the feasibility of procedural steps and the correctness of answers. This improvement addresses the limitations of previous methods, which often relied on simple binary conclusions without providing reasoning behind the judgments.
\subsubsection{Step Evaluator}
When each node is generated, the Step Evaluator assesses potential success of the steps taken in the problem-solving process through a multi-dimensional approach:

\noindent\textbf{Comprehensive Step Analysis:} The evaluator reviews the entire trajectory $z_{i1,\ldots, ij}$, including the thoughts process, actions taken, and observations made.

\noindent\textbf{Feasibility Assessment:} It evaluates the likelihood that the sequence of steps will lead to the correct answer. This process can be formalized as evaluating the value function \( V(p, z_{i1,\ldots, ij})(z_{ij}) \) for the sequence of steps $z_{i1,\ldots, ij}$, where the current step $z_{ij}$ is assessed according to the likelihood \( p(v \mid z_{ij}) \) of leading to a successful outcome:
\[
V(p, z_{i1,\ldots, ij})(z_{ij}) \sim p(v \mid z_{ij}).
\]

\noindent\textbf{Detailed Justification:} Each judgment includes an analysis, explaining the likelihood of a step leading to the correct answer, enhancing the transparency and value of the model's output.

\noindent\textbf{Categorization:} Based on comprehensive assessment, the feasibility of steps is categorized into levels "sure," "maybe," or "impossible", which is the scalar value $v$. Steps are then explored in order of priority, focusing on the most promising paths first.

\subsubsection{Answer Evaluator}

\label{subsubsec:answer_evaluator}
The Answer Evaluator leverages the capabilities of large language models to assess the relevance of generated result in relation to the original query. Unlike traditional methods that focus solely on correctness, which may result in a misalignment between the question and its generated answer (e.g. asking about the championship-winning team but receiving a response with the name of an individual from that team), the evaluator examines whether the generated result is on-topic by considering its alignment with the context and intent of the query. This process involves presenting the LLM with both the query and the generated result, prompting it to evaluate whether the generated result solves the problem and whether it is on-topic, and gives the reasons for the judgment. The integration of detailed reasoning in the Answer Evaluator promotes a deeper understanding of the model’s decisions, significantly improving the quality and reliability of the outputs.

\subsection{Explore Algorithm}
The Explore Algorithm in our framework employs a depth-first search strategy to navigate through the solution space. This approach allows the algorithm to dive deep into each reasoning node, continuing until it reaches an endpoint or meets specific conditions, which aligns perfectly with our framework of constructing and verifying a complete reasoning chain.
Additionally, the DFS strategy's deep exploration creates paths that are ideal for monitoring and adjustment by our Answer Evaluator and Error Analysis. If the Answer Evaluator identifies a path as invalid, the algorithm backtracks to explore other potential pathways. This systematic approach enables real-time identification and correction of erroneous paths, preventing the model from repeating the same mistakes and ensuring a more comprehensive and optimized exploration of the solution space. Figure~\ref{fig:mylabel} provides comparison between ReAct paradigm and FGDIP.

\begin{table*}
\centering
\begin{tabular}{l>{\centering\arraybackslash}p{2cm}>{\centering\arraybackslash}p{2cm}>{\centering\arraybackslash}p{2cm}>{\centering\arraybackslash}p{2cm}}
\hline
\multirow{2}{*}{\centering\textbf{Method}} & \multicolumn{4}{c}{\textbf{HotpotQA}} \\
\cline{2-5}
 & Easy & Medium & Hard & Average \\
\hline
Standard \citep{brown2020language} & 52.46 & 34.06 & 30.55 & 39.33 \\
CoT \citep{wei2023chainofthought} & 48.64 & 49.96 & 36.32 & 45.16 \\
ReAct \citep{yao2022react} & 44.53 & 39.44 & 36.23 & 40.18 \\
Reflexion \citep{shinn2023reflexion} & 52.61 & 50.83 & \underline{39.90} & 47.96 \\
Dr3$^*$ \citep{gao2024dr3} & - & - & - & 46.53\\
UALA \citep{han2024uncertaintyawarelanguageagent} & \underline{57.38} & \underline{52.07} & 38.02 & \underline{49.44} \\
FGDIP (Ours) & \textbf{60.46} & \textbf{53.87} & \textbf{48.56} & \textbf{54.47} \\
\hline
\end{tabular}

\caption{\label{citation-guide}
Main results of FGDIP compared to various baselines on HotpotQA dataset. The evaluation metrics are F1 Score (\%). Bold shows the best result for each column and underline shows the second-best results for each column. Results denoted by $^*$ are reported from \citet{gao2024dr3}.}
\end{table*}

\section{Experiments}
\subsection{Experimental Setting}

\textbf{Tasks.} We evaluate our model on HotpotQA \citep{yang2018hotpotqa} and StrategyQA \citep{10.1162/tacl_a_00370} datasets. HotpotQA requires multi-hop reasoning over Wikipedia passages with free-form answers. Following \citet{qiao2024autoact}, we use their datasets which consist of questions divided into three levels for evaluation. StrategyQA is an open-domain dataset with binary answers, focusing on implicit reasoning where queries don't explicitly guide the derivation of correct answers. We use the same dev set used in FireAct \citep{chen2023fireact} to evaluate our model.

\noindent\textbf{Baselines.} We use GPT-3.5-Turbo, GPT-4o-mini, GPT-4o \citep{openai2024gpt4}, and Gemini-1.5-flash \citep{team2023gemini} as the inference engines. We compare FGDIP with 1) \textbf{Standard} \citep{brown2020language}, a method utilizing the basic capabilities of model for inference with a few example prompts. 2) \textbf{Chain-of-Thought} \citep{wei2023chainofthought}, a reasoning approach that mimics human thought patterns, allowing the model to explain its reasoning steps. 3) \textbf{ReAct} \citep{yao2022react}, a paradigm that synergizes reasoning and action, enabling the model to tackle complex tasks by adapting to real-time interactions. 4) \textbf{Reflexion} \citep{shinn2023reflexion}, applied to ReAct, incorporating ground truth label feedback to iteratively enhance performance by enabling the model to reflect on its previous outputs. 5) \textbf{Dr3} \citep{gao2024dr3}, a mechanism that detects and corrects off-topic answers in large language models through iterative reasoning and backtracking. 6) \textbf{UALA} \citep{han2024uncertaintyawarelanguageagent}, a language agent framework that manages interaction uncertainty to optimize reasoning and reduce external dependencies. We evaluate the performance of each method using the F1 score, which measures the agreement between the predicted and ground-truth answers, with the reward ranging from 0 to 1.

\begin{table}[!htbp]
\centering
\begin{tabular}{lcc}
\hline
\textbf{Method} & \textbf{StrategyQA}  \\
\hline
Standard \citep{brown2020language} & 61.35\\
CoT \citep{wei2023chainofthought} &34.30\\
ReAct \citep{yao2022react}  &49.76\\
Reflexion \citep{shinn2023reflexion}  &56.52\\
UALA \citep{han2024uncertaintyawarelanguageagent}  &\underline{62.80}\\
FGDIP (Ours)  &\textbf{70.05}\\

\hline
\end{tabular}
\caption{Main results of FGDIP compared to various baselines on StrategyQA. The evaluation metrics are F1 Score (\%). Bold shows the best result for each column and underline shows the second-best results for each column.}
\label{strategyqa}
\end{table}

\subsection{Main Results}
Table~\ref{citation-guide} and ~\ref{strategyqa} detail FGDIP's performance against various baselines using GPT-3.5-Turbo, while Appendix~\ref{model_generalizability} demonstrates FGDIP's generalizability across different large language models.

\noindent\textbf{HotpotQA Datasets Analysis.} As shown in Table~\ref{citation-guide}, FGDIP's F1 scores across different difficulty levels are notably higher than the baselines. In particular, FGDIP achieves a 60.46\% F1 score for easy questions, a 53.87\% F1 score for medium questions, and a 48.56\% F1 score for hard questions. These results represent substantial improvements of 3.08\%, 1.80\%, and 8.66\% respectively compared to the best baseline \citep{han2024uncertaintyawarelanguageagent, shinn2023reflexion}. The primary reason for this improvement is FGDIP's ability to dynamically adjust its reasoning process by integrating real-time feedback and historical error analysis, which allows it to navigate complex, multi-hop questions more effectively than static methods like Chain-of-Thought.

\begin{figure*}[h]
\centering
\begin{minipage}[b]{0.32\textwidth}
  \centering
  \includegraphics[height=0.12\textheight, keepaspectratio]{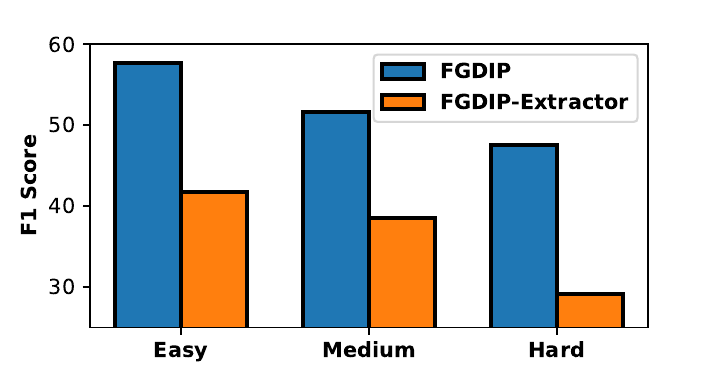} 
  \subcaption{\footnotesize  Without Extractor}
  \label{fig:image1}
\end{minipage}
\hfill 
\begin{minipage}[b]{0.32\textwidth}
  \centering
  \includegraphics[height=0.12\textheight, keepaspectratio]{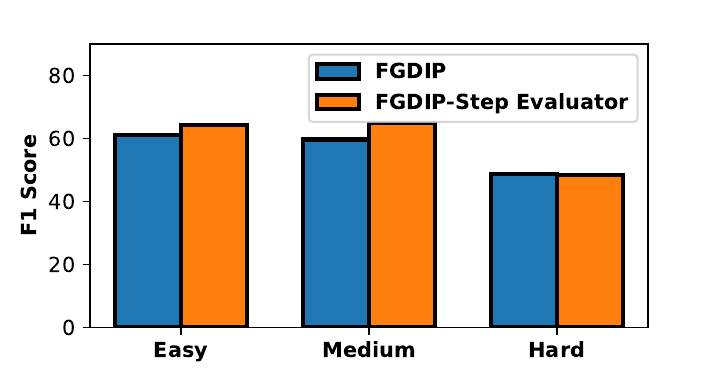} 
  \subcaption{\footnotesize  Without Step Evaluator}
  \label{fig:image2}
\end{minipage}
\hfill 
\begin{minipage}[b]{0.32\textwidth}
  \centering
  \includegraphics[height=0.12\textheight, keepaspectratio]{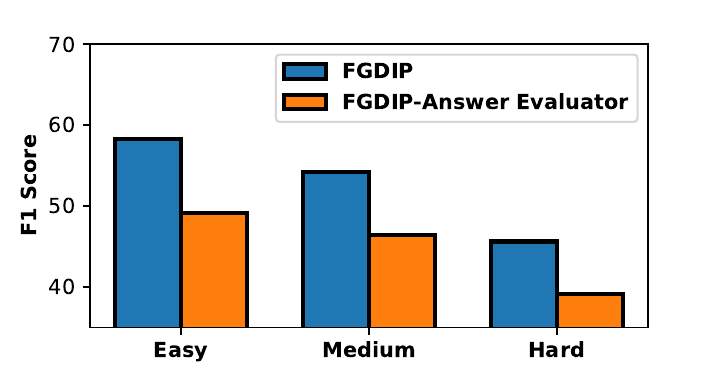} 
  \subcaption{\footnotesize  Without Answer Evaluator}
  \label{fig:image3}
\end{minipage}

\caption{Performance analysis of model components on F1 Score: (a) Framework without the Extractor module, illustrating the impact of omitting the component responsible for breaking down complex queries, (b) Framework without the Step Evaluator, showing effective in maintaining performance while minimizing resource consumption, and (c) Framework without the Answer Evaluator, showing the outcomes of excluding the module for iterative answer validation and correction.}
\label{fig:system_components}
\end{figure*}

\noindent\textbf{StrategyQA Datasets Analysis.} In the StrategyQA dataset, as shown in Table~\ref{strategyqa}, FGDIP achieves an F1 score of 70.05\%, outperforming the best baseline UALA \citep{han2024uncertaintyawarelanguageagent}, which had an F1 score of 62.80\%. The 7.25\% improvement is due to FGDIP's iterative approach that breaks down complex reasoning tasks and refines strategies based on real-time feedback and past errors. This is crucial for StrategyQA, where indirect reasoning challenges static methods.

\noindent\textbf{Performance Breakdown and Reasons.} The detailed performance breakdown reveals that FGDIP’s advantage grows with the complexity of the questions. For difficult questions that require multiple reasoning steps, the dynamic adaptation of FGDIP allows it to explore a broader search space and identify more relevant information paths, leading to more accurate answers. This is particularly evident in HotpotQA's hard question category, where FGDIP outperforms the baseline by 8.66\%.

In contrast, for simpler questions where the reasoning path is straightforward, the performance gap between FGDIP and the baselines is narrower. This indicates that the benefits of FGDIP's dynamic and adaptive strategy are maximized in scenarios where complex reasoning and flexible information exploration are required.

\section{Analysis}

In this section, we analyze the effectiveness and necessity of each module within FGDIP. We conduct ablation studies using HotpotQA datasets. The influence of each module on the F1 Score performance is illustrated in the Figure~\ref{fig:system_components}. Additionally, we explore the framework's applicability to closed-domain problems and conduct cost analysis to quantify the trade-offs involved in achieving improved reasoning capabilities.

\subsection{Multivariate Information Extractor}

The removal of the Extractor module from our model results in significant declines in both F1 scores and accuracy across various difficulty levels, as detailed in Appendix~\ref{Extractor Module}. 

As shown in Table~\ref{tab:rates1}, accuracy declines by 13.16\% on easy tasks, 12.68\% on medium tasks, and 18.42\% on hard tasks, with F1 scores exhibiting similar declines. This sharp decrease highlights that the Extractor module is essential for identifying and integrating relevant information flexibly. Without it, the model adopts a rigid, less effective approach, unable to refocus based on the evolving information landscape, particularly in complex scenarios with multiple interrelated entities. This study underscores the Extractor module’s critical role in avoiding the pitfalls of conventional linear methods by dynamically extracting key information, which enhances the model’s adaptability and accuracy in handling complex, diverse problems.

\subsection{Step Evaluator}

We assess the impact of the Step Evaluator by comparing model performance with and without this module, as accuracy shown in Appendix~\ref{Step Evaluator Module}. Removing the Step Evaluator leads to slight accuracy gains on easy and medium tasks, increasing by 3.08\% and 5.66\%, respectively, while decreasing by 1.69\% on hard tasks. In theory, removing this module might allow more freedom in exploring solutions and potentially expand the search scope. However, the modest improvement in performance confirms our initial hypothesis: the Step Evaluator is highly effective in maintaining performance while minimizing resource consumption, such as reducing token usage and processing time.

While removing the Step Evaluator offers minor gains on simpler tasks, it reduces efficiency and increases resource use. Without it, the model tends to stray into irrelevant information on complex tasks, lowering performance on intricate reasoning steps. This highlights the Step Evaluator’s role in guiding the model effectively and maintaining accuracy across varying task complexities.

\subsection{Answer Evaluator and Error Analysis}
In this ablation study, we evaluate the impact of removing the Answer Evaluator and Error Analysis from our model, with the detailed results available in Appendix~\ref{Answer Evaluator Module}. Without these modules, the model's process concludes immediately after generating an initial answer, forgoing any further validation or correction.

F1 scores drop by 9.10\% on easy, 7.80\% on medium, and 6.44\% on hard tasks, with similar declines in accuracy. This decline underscores that the components not only help the model verify the accuracy and relevance of its answers but also facilitate continuous refinement of its reasoning process, leading to more precise and appropriate solutions.
Without these modules, the model's outputs became less reliable, particularly in tasks requiring deeper reasoning. These components are essential for refining answers and ensuring that mistakes are addressed iteratively, highlighting their fundamental importance in sustaining the model's performance on complex tasks. Their absence underscores the critical role that continuous evaluation and correction play in achieving accurate results.

\subsection{Applicability to mathematical reasoning Problems}
In addition to its effectiveness in open-domain multi-hop reasoning tasks, our FGDIP framework also demonstrates strong applicability to closed-domain problems. Specifically, we evaluate the performance of FGDIP on a mathematical reasoning task using the Game of 24 dataset, a benchmark for assessing logical and numerical reasoning capabilities, where we use GPT-4 as the inference engine.

As shown in the Table~\ref{tab:game}, FGDIP, with only $k\leq 3$, achieves an accuracy of 70.73\%, surpassing other methods such as Standard, CoT, and ToT. This superior performance highlights the framework's versatility and robustness, as it can adapt its dynamic reasoning strategies to handle not only the open-domain tasks but also the structured and rule-based challenges presented by closed-domain problems like mathematical reasoning. 

\begin{table}[h]
\centering
\begin{tabular}{lc}
\hline
\textbf{Method} & \textbf{Acc} \\
\hline
Standard $(k=100)$ & 36.59\%  \\
CoT $(k=100)$ & 48.78\%  \\
ToT $(k=5)$ & 68.29 \% \\
FGDIP $(k\leq 3)$& \textbf{70.73\%}\\
\hline
\end{tabular}
\caption{Accuracy comparison on the Game of 24, where k represents best of k samples. We use GPT-4 as the inference engine.}
\label{tab:game}
\end{table}
\subsection{Cost Analysis}
FGDIP's dynamic adjustments and real-time feedback enhance accuracy but increase costs. To quantify impact, we conduct an analysis of the token and time costs associated with various methods.

As shown in Table~\ref{tab:token cost}, FGDIP only represents a fourfold increase over Reflexion. This contrast is modest compared to the token consumption leap from CoT \citep{wei2023chainofthought} to ReAct \citep{yao2022react}, which sees nearly a thirtyfold increase. Despite FGDIP's higher token count, the additional resource use is justified by FGDIP's dynamic search and refinement, crucial for complex reasoning.

In terms of time cost, as shown in Table~\ref{tab:costs}, FGDIP also incurs a higher processing time compared to other methods. Despite these increases, the improvements in accuracy and reasoning strategy make these costs feasible and valuable for solving open-domain problems, where extensive exploration and real-time adjustments are crucial.

\section{Conclusion}
\label{sec:bibtex}
This study introduces a novel framework for enhancing language agents in open-domain multi-hop reasoning tasks. Using dynamic extraction and adaptive strategies, our approach overcomes traditional fixed-sequence limitations. Experimental results demonstrate significant improvements in accuracy and efficiency, particularly on complex datasets like HotpotQA and StrategyQA.

Ablation studies confirm the vital roles of FGDIP’s components. The Extractor module efficiently manages and processes complex information, while the Step Evaluator strategically prioritizes the most promising paths, thereby enhancing overall efficiency and decision-making. The Answer Evaluator and Error Analysis ensure accuracy through continuous refinement. Moreover, the framework’s effectiveness extends to closed-domain problems, where it demonstrates adaptability and robust performance, proving its versatility across varied problems.

Our framework significantly advances language agents by offering a robust, adaptable approach that enhances multi-hop reasoning in complex scenarios.

\section*{Limitations}
FGDIP’s dynamic adjustments and real-time feedback lead to high costs due to frequent calls to large language models, especially in large-scale open-domain problems, increasing processing times and resource use. The evaluation mechanism, while detailed, relies on predefined rules that may not cover all error types, affecting accuracy and relevance.
\section*{Acknowledgements}
This work was supported by the Beijing Natural Science Foundation (IS23059).

\bibliography{main}

\begin{thebibliography}{24}
\providecommand{\natexlab}[1]{#1}

\bibitem[{Besta et~al.(2024)Besta, Blach, Kubicek, Gerstenberger, Podstawski, Gianinazzi, Gajda, Lehmann, Niewiadomski, Nyczyk, and Hoefler}]{Besta_2024}
Maciej Besta, Nils Blach, Ales Kubicek, Robert Gerstenberger, Michal Podstawski, Lukas Gianinazzi, Joanna Gajda, Tomasz Lehmann, Hubert Niewiadomski, Piotr Nyczyk, and Torsten Hoefler. 2024.
\newblock \href {https://doi.org/10.1609/aaai.v38i16.29720} {Graph of thoughts: Solving elaborate problems with large language models}.
\newblock \emph{Proceedings of the AAAI Conference on Artificial Intelligence}, 38(16):17682–17690.

\bibitem[{Brown et~al.(2020)Brown, Mann, Ryder, Subbiah, Kaplan, Dhariwal, Neelakantan, Shyam, Sastry, Askell et~al.}]{brown2020language}
Tom Brown, Benjamin Mann, Nick Ryder, Melanie Subbiah, Jared~D Kaplan, Prafulla Dhariwal, Arvind Neelakantan, Pranav Shyam, Girish Sastry, Amanda Askell, et~al. 2020.
\newblock Language models are few-shot learners.
\newblock \emph{Advances in neural information processing systems}, 33:1877--1901.

\bibitem[{Chen et~al.(2023{\natexlab{a}})Chen, Shu, Shareghi, Collier, Narasimhan, and Yao}]{chen2023fireact}
Baian Chen, Chang Shu, Ehsan Shareghi, Nigel Collier, Karthik Narasimhan, and Shunyu Yao. 2023{\natexlab{a}}.
\newblock \href {https://arxiv.org/abs/2310.05915} {Fireact: Toward language agent fine-tuning}.
\newblock \emph{Preprint}, arXiv:2310.05915.

\bibitem[{Chen et~al.(2023{\natexlab{b}})Chen, Su, Zuo, Yang, Yuan, Chan, Yu, Lu, Hung, Qian, Qin, Cong, Xie, Liu, Sun, and Zhou}]{chen2023agentverse}
Weize Chen, Yusheng Su, Jingwei Zuo, Cheng Yang, Chenfei Yuan, Chi-Min Chan, Heyang Yu, Yaxi Lu, Yi-Hsin Hung, Chen Qian, Yujia Qin, Xin Cong, Ruobing Xie, Zhiyuan Liu, Maosong Sun, and Jie Zhou. 2023{\natexlab{b}}.
\newblock \href {https://arxiv.org/abs/2308.10848} {Agentverse: Facilitating multi-agent collaboration and exploring emergent behaviors}.
\newblock \emph{Preprint}, arXiv:2308.10848.

\bibitem[{Gao et~al.(2024)Gao, Zhu, Cao, Zhou, Wu, Chen, Wu, Hu, and Dai}]{gao2024dr3}
Yuan Gao, Yiheng Zhu, Yuanbin Cao, Yinzhi Zhou, Zhen Wu, Yujie Chen, Shenglan Wu, Haoyuan Hu, and Xinyu Dai. 2024.
\newblock \href {https://arxiv.org/abs/2403.12393} {Dr3: Ask large language models not to give off-topic answers in open domain multi-hop question answering}.
\newblock \emph{Preprint}, arXiv:2403.12393.

\bibitem[{Geva et~al.(2021)Geva, Khashabi, Segal, Khot, Roth, and Berant}]{10.1162/tacl_a_00370}
Mor Geva, Daniel Khashabi, Elad Segal, Tushar Khot, Dan Roth, and Jonathan Berant. 2021.
\newblock \href {https://doi.org/10.1162/tacl_a_00370} {{Did Aristotle Use a Laptop? A Question Answering Benchmark with Implicit Reasoning Strategies}}.
\newblock \emph{Transactions of the Association for Computational Linguistics}, 9:346--361.

\bibitem[{Han et~al.(2024)Han, Buntine, and Shareghi}]{han2024uncertaintyawarelanguageagent}
Jiuzhou Han, Wray Buntine, and Ehsan Shareghi. 2024.
\newblock \href {https://arxiv.org/abs/2401.14016} {Towards uncertainty-aware language agent}.
\newblock \emph{Preprint}, arXiv:2401.14016.

\bibitem[{Hong et~al.(2023)Hong, Zhuge, Chen, Zheng, Cheng, Zhang, Wang, Wang, Yau, Lin, Zhou, Ran, Xiao, Wu, and Schmidhuber}]{hong2023metagpt}
Sirui Hong, Mingchen Zhuge, Jonathan Chen, Xiawu Zheng, Yuheng Cheng, Ceyao Zhang, Jinlin Wang, Zili Wang, Steven Ka~Shing Yau, Zijuan Lin, Liyang Zhou, Chenyu Ran, Lingfeng Xiao, Chenglin Wu, and Jürgen Schmidhuber. 2023.
\newblock \href {https://arxiv.org/abs/2308.00352} {Metagpt: Meta programming for a multi-agent collaborative framework}.
\newblock \emph{Preprint}, arXiv:2308.00352.

\bibitem[{Lu et~al.(2023)Lu, Peng, Cheng, Galley, Chang, Wu, Zhu, and Gao}]{lu2023chameleon}
Pan Lu, Baolin Peng, Hao Cheng, Michel Galley, Kai-Wei Chang, Ying~Nian Wu, Song-Chun Zhu, and Jianfeng Gao. 2023.
\newblock \href {https://arxiv.org/abs/2304.09842} {Chameleon: Plug-and-play compositional reasoning with large language models}.
\newblock \emph{Preprint}, arXiv:2304.09842.

\bibitem[{Madaan et~al.(2023)Madaan, Tandon, Gupta, Hallinan, Gao, Wiegreffe, Alon, Dziri, Prabhumoye, Yang, Gupta, Majumder, Hermann, Welleck, Yazdanbakhsh, and Clark}]{madaan2023selfrefine}
Aman Madaan, Niket Tandon, Prakhar Gupta, Skyler Hallinan, Luyu Gao, Sarah Wiegreffe, Uri Alon, Nouha Dziri, Shrimai Prabhumoye, Yiming Yang, Shashank Gupta, Bodhisattwa~Prasad Majumder, Katherine Hermann, Sean Welleck, Amir Yazdanbakhsh, and Peter Clark. 2023.
\newblock \href {https://arxiv.org/abs/2303.17651} {Self-refine: Iterative refinement with self-feedback}.
\newblock \emph{Preprint}, arXiv:2303.17651.

\bibitem[{OpenAI et~al.(2024)OpenAI, Achiam, Adler, Agarwal, Ahmad, Akkaya, Aleman, Almeida, Altenschmidt, Altman, Anadkat, Avila, Babuschkin, Balaji, Balcom, Baltescu, Bao, Bavarian, Belgum, Bello, Berdine, Bernadett-Shapiro, Berner, Bogdonoff, Boiko, Boyd, Brakman, Brockman, Brooks, Brundage, Button, Cai, Campbell, Cann, Carey, Carlson, Carmichael, Chan, Chang, Chantzis, Chen, Chen, Chen, Chen, Chen, Chess, Cho, Chu, Chung, Cummings, Currier, Dai, Decareaux, Degry, Deutsch, Deville, Dhar, Dohan, Dowling, Dunning, Ecoffet, Eleti, Eloundou, Farhi, Fedus, Felix, Fishman, Forte, Fulford, Gao, Georges, Gibson, Goel, Gogineni, Goh, Gontijo-Lopes, Gordon, Grafstein, Gray, Greene, Gross, Gu, Guo, Hallacy, Han, Harris, He, Heaton, Heidecke, Hesse, Hickey, Hickey, Hoeschele, Houghton, Hsu, Hu, Hu, Huizinga, Jain, Jain, Jang, Jiang, Jiang, Jin, Jin, Jomoto, Jonn, Jun, Kaftan, Łukasz Kaiser, Kamali, Kanitscheider, Keskar, Khan, Kilpatrick, Kim, Kim, Kim, Kirchner, Kiros, Knight, Kokotajlo, Łukasz Kondraciuk,
  Kondrich, Konstantinidis, Kosic, Krueger, Kuo, Lampe, Lan, Lee, Leike, Leung, Levy, Li, Lim, Lin, Lin, Litwin, Lopez, Lowe, Lue, Makanju, Malfacini, Manning, Markov, Markovski, Martin, Mayer, Mayne, McGrew, McKinney, McLeavey, McMillan, McNeil, Medina, Mehta, Menick, Metz, Mishchenko, Mishkin, Monaco, Morikawa, Mossing, Mu, Murati, Murk, Mély, Nair, Nakano, Nayak, Neelakantan, Ngo, Noh, Ouyang, O'Keefe, Pachocki, Paino, Palermo, Pantuliano, Parascandolo, Parish, Parparita, Passos, Pavlov, Peng, Perelman, de~Avila Belbute~Peres, Petrov, de~Oliveira~Pinto, Michael, Pokorny, Pokrass, Pong, Powell, Power, Power, Proehl, Puri, Radford, Rae, Ramesh, Raymond, Real, Rimbach, Ross, Rotsted, Roussez, Ryder, Saltarelli, Sanders, Santurkar, Sastry, Schmidt, Schnurr, Schulman, Selsam, Sheppard, Sherbakov, Shieh, Shoker, Shyam, Sidor, Sigler, Simens, Sitkin, Slama, Sohl, Sokolowsky, Song, Staudacher, Such, Summers, Sutskever, Tang, Tezak, Thompson, Tillet, Tootoonchian, Tseng, Tuggle, Turley, Tworek, Uribe, Vallone,
  Vijayvergiya, Voss, Wainwright, Wang, Wang, Wang, Ward, Wei, Weinmann, Welihinda, Welinder, Weng, Weng, Wiethoff, Willner, Winter, Wolrich, Wong, Workman, Wu, Wu, Wu, Xiao, Xu, Yoo, Yu, Yuan, Zaremba, Zellers, Zhang, Zhang, Zhao, Zheng, Zhuang, Zhuk, and Zoph}]{openai2024gpt4}
OpenAI, Josh Achiam, Steven Adler, Sandhini Agarwal, Lama Ahmad, Ilge Akkaya, Florencia~Leoni Aleman, Diogo Almeida, Janko Altenschmidt, Sam Altman, Shyamal Anadkat, Red Avila, Igor Babuschkin, Suchir Balaji, Valerie Balcom, Paul Baltescu, Haiming Bao, Mohammad Bavarian, Jeff Belgum, Irwan Bello, Jake Berdine, Gabriel Bernadett-Shapiro, Christopher Berner, Lenny Bogdonoff, Oleg Boiko, Madelaine Boyd, Anna-Luisa Brakman, Greg Brockman, Tim Brooks, Miles Brundage, Kevin Button, Trevor Cai, Rosie Campbell, Andrew Cann, Brittany Carey, Chelsea Carlson, Rory Carmichael, Brooke Chan, Che Chang, Fotis Chantzis, Derek Chen, Sully Chen, Ruby Chen, Jason Chen, Mark Chen, Ben Chess, Chester Cho, Casey Chu, Hyung~Won Chung, Dave Cummings, Jeremiah Currier, Yunxing Dai, Cory Decareaux, Thomas Degry, Noah Deutsch, Damien Deville, Arka Dhar, David Dohan, Steve Dowling, Sheila Dunning, Adrien Ecoffet, Atty Eleti, Tyna Eloundou, David Farhi, Liam Fedus, Niko Felix, Simón~Posada Fishman, Juston Forte, Isabella Fulford, Leo
  Gao, Elie Georges, Christian Gibson, Vik Goel, Tarun Gogineni, Gabriel Goh, Rapha Gontijo-Lopes, Jonathan Gordon, Morgan Grafstein, Scott Gray, Ryan Greene, Joshua Gross, Shixiang~Shane Gu, Yufei Guo, Chris Hallacy, Jesse Han, Jeff Harris, Yuchen He, Mike Heaton, Johannes Heidecke, Chris Hesse, Alan Hickey, Wade Hickey, Peter Hoeschele, Brandon Houghton, Kenny Hsu, Shengli Hu, Xin Hu, Joost Huizinga, Shantanu Jain, Shawn Jain, Joanne Jang, Angela Jiang, Roger Jiang, Haozhun Jin, Denny Jin, Shino Jomoto, Billie Jonn, Heewoo Jun, Tomer Kaftan, Łukasz Kaiser, Ali Kamali, Ingmar Kanitscheider, Nitish~Shirish Keskar, Tabarak Khan, Logan Kilpatrick, Jong~Wook Kim, Christina Kim, Yongjik Kim, Jan~Hendrik Kirchner, Jamie Kiros, Matt Knight, Daniel Kokotajlo, Łukasz Kondraciuk, Andrew Kondrich, Aris Konstantinidis, Kyle Kosic, Gretchen Krueger, Vishal Kuo, Michael Lampe, Ikai Lan, Teddy Lee, Jan Leike, Jade Leung, Daniel Levy, Chak~Ming Li, Rachel Lim, Molly Lin, Stephanie Lin, Mateusz Litwin, Theresa Lopez, Ryan
  Lowe, Patricia Lue, Anna Makanju, Kim Malfacini, Sam Manning, Todor Markov, Yaniv Markovski, Bianca Martin, Katie Mayer, Andrew Mayne, Bob McGrew, Scott~Mayer McKinney, Christine McLeavey, Paul McMillan, Jake McNeil, David Medina, Aalok Mehta, Jacob Menick, Luke Metz, Andrey Mishchenko, Pamela Mishkin, Vinnie Monaco, Evan Morikawa, Daniel Mossing, Tong Mu, Mira Murati, Oleg Murk, David Mély, Ashvin Nair, Reiichiro Nakano, Rajeev Nayak, Arvind Neelakantan, Richard Ngo, Hyeonwoo Noh, Long Ouyang, Cullen O'Keefe, Jakub Pachocki, Alex Paino, Joe Palermo, Ashley Pantuliano, Giambattista Parascandolo, Joel Parish, Emy Parparita, Alex Passos, Mikhail Pavlov, Andrew Peng, Adam Perelman, Filipe de~Avila Belbute~Peres, Michael Petrov, Henrique~Ponde de~Oliveira~Pinto, Michael, Pokorny, Michelle Pokrass, Vitchyr~H. Pong, Tolly Powell, Alethea Power, Boris Power, Elizabeth Proehl, Raul Puri, Alec Radford, Jack Rae, Aditya Ramesh, Cameron Raymond, Francis Real, Kendra Rimbach, Carl Ross, Bob Rotsted, Henri Roussez,
  Nick Ryder, Mario Saltarelli, Ted Sanders, Shibani Santurkar, Girish Sastry, Heather Schmidt, David Schnurr, John Schulman, Daniel Selsam, Kyla Sheppard, Toki Sherbakov, Jessica Shieh, Sarah Shoker, Pranav Shyam, Szymon Sidor, Eric Sigler, Maddie Simens, Jordan Sitkin, Katarina Slama, Ian Sohl, Benjamin Sokolowsky, Yang Song, Natalie Staudacher, Felipe~Petroski Such, Natalie Summers, Ilya Sutskever, Jie Tang, Nikolas Tezak, Madeleine~B. Thompson, Phil Tillet, Amin Tootoonchian, Elizabeth Tseng, Preston Tuggle, Nick Turley, Jerry Tworek, Juan Felipe~Cerón Uribe, Andrea Vallone, Arun Vijayvergiya, Chelsea Voss, Carroll Wainwright, Justin~Jay Wang, Alvin Wang, Ben Wang, Jonathan Ward, Jason Wei, CJ~Weinmann, Akila Welihinda, Peter Welinder, Jiayi Weng, Lilian Weng, Matt Wiethoff, Dave Willner, Clemens Winter, Samuel Wolrich, Hannah Wong, Lauren Workman, Sherwin Wu, Jeff Wu, Michael Wu, Kai Xiao, Tao Xu, Sarah Yoo, Kevin Yu, Qiming Yuan, Wojciech Zaremba, Rowan Zellers, Chong Zhang, Marvin Zhang, Shengjia
  Zhao, Tianhao Zheng, Juntang Zhuang, William Zhuk, and Barret Zoph. 2024.
\newblock \href {https://arxiv.org/abs/2303.08774} {Gpt-4 technical report}.
\newblock \emph{Preprint}, arXiv:2303.08774.

\bibitem[{Qiao et~al.(2023)Qiao, Ou, Zhang, Chen, Yao, Deng, Tan, Huang, and Chen}]{qiao2023reasoning}
Shuofei Qiao, Yixin Ou, Ningyu Zhang, Xiang Chen, Yunzhi Yao, Shumin Deng, Chuanqi Tan, Fei Huang, and Huajun Chen. 2023.
\newblock \href {https://arxiv.org/abs/2212.09597} {Reasoning with language model prompting: A survey}.
\newblock \emph{Preprint}, arXiv:2212.09597.

\bibitem[{Qiao et~al.(2024)Qiao, Zhang, Fang, Luo, Zhou, Jiang, Lv, and Chen}]{qiao2024autoact}
Shuofei Qiao, Ningyu Zhang, Runnan Fang, Yujie Luo, Wangchunshu Zhou, Yuchen~Eleanor Jiang, Chengfei Lv, and Huajun Chen. 2024.
\newblock \href {https://arxiv.org/abs/2401.05268} {Autoact: Automatic agent learning from scratch via self-planning}.
\newblock \emph{Preprint}, arXiv:2401.05268.

\bibitem[{Qin et~al.(2023)Qin, Liang, Ye, Zhu, Yan, Lu, Lin, Cong, Tang, Qian, Zhao, Hong, Tian, Xie, Zhou, Gerstein, Li, Liu, and Sun}]{qin2023toolllm}
Yujia Qin, Shihao Liang, Yining Ye, Kunlun Zhu, Lan Yan, Yaxi Lu, Yankai Lin, Xin Cong, Xiangru Tang, Bill Qian, Sihan Zhao, Lauren Hong, Runchu Tian, Ruobing Xie, Jie Zhou, Mark Gerstein, Dahai Li, Zhiyuan Liu, and Maosong Sun. 2023.
\newblock \href {https://arxiv.org/abs/2307.16789} {Toolllm: Facilitating large language models to master 16000+ real-world apis}.
\newblock \emph{Preprint}, arXiv:2307.16789.

\bibitem[{Shinn et~al.(2023)Shinn, Cassano, Berman, Gopinath, Narasimhan, and Yao}]{shinn2023reflexion}
Noah Shinn, Federico Cassano, Edward Berman, Ashwin Gopinath, Karthik Narasimhan, and Shunyu Yao. 2023.
\newblock \href {https://arxiv.org/abs/2303.11366} {Reflexion: Language agents with verbal reinforcement learning}.
\newblock \emph{Preprint}, arXiv:2303.11366.

\bibitem[{Team et~al.(2023)Team, Anil, Borgeaud, Alayrac, Yu, Soricut, Schalkwyk, Dai, Hauth, Millican et~al.}]{team2023gemini}
Gemini Team, Rohan Anil, Sebastian Borgeaud, Jean-Baptiste Alayrac, Jiahui Yu, Radu Soricut, Johan Schalkwyk, Andrew~M Dai, Anja Hauth, Katie Millican, et~al. 2023.
\newblock Gemini: a family of highly capable multimodal models.
\newblock \emph{arXiv preprint arXiv:2312.11805}.

\bibitem[{Wang et~al.(2024)Wang, Ma, Feng, Zhang, Yang, Zhang, Chen, Tang, Chen, Lin, Zhao, Wei, and Wen}]{Wang_2024}
Lei Wang, Chen Ma, Xueyang Feng, Zeyu Zhang, Hao Yang, Jingsen Zhang, Zhiyuan Chen, Jiakai Tang, Xu~Chen, Yankai Lin, Wayne~Xin Zhao, Zhewei Wei, and Jirong Wen. 2024.
\newblock \href {https://doi.org/10.1007/s11704-024-40231-1} {A survey on large language model based autonomous agents}.
\newblock \emph{Frontiers of Computer Science}, 18(6).

\bibitem[{Wang et~al.(2023)Wang, Wei, Schuurmans, Le, Chi, Narang, Chowdhery, and Zhou}]{wang2023selfconsistency}
Xuezhi Wang, Jason Wei, Dale Schuurmans, Quoc Le, Ed~Chi, Sharan Narang, Aakanksha Chowdhery, and Denny Zhou. 2023.
\newblock \href {https://arxiv.org/abs/2203.11171} {Self-consistency improves chain of thought reasoning in language models}.
\newblock \emph{Preprint}, arXiv:2203.11171.

\bibitem[{Wei et~al.(2023)Wei, Wang, Schuurmans, Bosma, Ichter, Xia, Chi, Le, and Zhou}]{wei2023chainofthought}
Jason Wei, Xuezhi Wang, Dale Schuurmans, Maarten Bosma, Brian Ichter, Fei Xia, Ed~Chi, Quoc Le, and Denny Zhou. 2023.
\newblock \href {https://arxiv.org/abs/2201.11903} {Chain-of-thought prompting elicits reasoning in large language models}.
\newblock \emph{Preprint}, arXiv:2201.11903.

\bibitem[{Xi et~al.(2023)Xi, Chen, Guo, He, Ding, Hong, Zhang, Wang, Jin, Zhou, Zheng, Fan, Wang, Xiong, Zhou, Wang, Jiang, Zou, Liu, Yin, Dou, Weng, Cheng, Zhang, Qin, Zheng, Qiu, Huang, and Gui}]{xi2023rise}
Zhiheng Xi, Wenxiang Chen, Xin Guo, Wei He, Yiwen Ding, Boyang Hong, Ming Zhang, Junzhe Wang, Senjie Jin, Enyu Zhou, Rui Zheng, Xiaoran Fan, Xiao Wang, Limao Xiong, Yuhao Zhou, Weiran Wang, Changhao Jiang, Yicheng Zou, Xiangyang Liu, Zhangyue Yin, Shihan Dou, Rongxiang Weng, Wensen Cheng, Qi~Zhang, Wenjuan Qin, Yongyan Zheng, Xipeng Qiu, Xuanjing Huang, and Tao Gui. 2023.
\newblock \href {https://arxiv.org/abs/2309.07864} {The rise and potential of large language model based agents: A survey}.
\newblock \emph{Preprint}, arXiv:2309.07864.

\bibitem[{Yang et~al.(2018)Yang, Qi, Zhang, Bengio, Cohen, Salakhutdinov, and Manning}]{yang2018hotpotqa}
Zhilin Yang, Peng Qi, Saizheng Zhang, Yoshua Bengio, William~W. Cohen, Ruslan Salakhutdinov, and Christopher~D. Manning. 2018.
\newblock \href {https://arxiv.org/abs/1809.09600} {Hotpotqa: A dataset for diverse, explainable multi-hop question answering}.
\newblock \emph{Preprint}, arXiv:1809.09600.

\bibitem[{Yao et~al.(2023)Yao, Yu, Zhao, Shafran, Griffiths, Cao, and Narasimhan}]{yao2023tree}
Shunyu Yao, Dian Yu, Jeffrey Zhao, Izhak Shafran, Thomas~L. Griffiths, Yuan Cao, and Karthik Narasimhan. 2023.
\newblock \href {https://arxiv.org/abs/2305.10601} {Tree of thoughts: Deliberate problem solving with large language models}.
\newblock \emph{Preprint}, arXiv:2305.10601.

\bibitem[{Yao et~al.(2022)Yao, Zhao, Yu, Du, Shafran, Narasimhan, and Cao}]{yao2022react}
Shunyu Yao, Jeffrey Zhao, Dian Yu, Nan Du, Izhak Shafran, Karthik Narasimhan, and Yuan Cao. 2022.
\newblock React: Synergizing reasoning and acting in language models.
\newblock \emph{arXiv preprint arXiv:2210.03629}.

\bibitem[{Zhu et~al.(2024)Zhu, Qiao, Ou, Deng, Zhang, Lyu, Shen, Liang, Gu, and Chen}]{zhu2024knowagent}
Yuqi Zhu, Shuofei Qiao, Yixin Ou, Shumin Deng, Ningyu Zhang, Shiwei Lyu, Yue Shen, Lei Liang, Jinjie Gu, and Huajun Chen. 2024.
\newblock \href {https://arxiv.org/abs/2403.03101} {Knowagent: Knowledge-augmented planning for llm-based agents}.
\newblock \emph{Preprint}, arXiv:2403.03101.

\end{thebibliography}

\appendix

\newpage
\newgeometry{a4paper, margin=1in}

\appendix

\section{Experimental Configuration and Methodology Details}
\label{Experimental_Configuration}
\subsection{Language Model Configuration}
We employ GPT-3.5-Turbo, GPT-4o-mini, Gemini-1.5-flash, and GPT-4o  as our evaluation models with a sampling temperature of 0.7.

\subsection{Dataset Preprocessing}
All experimental datasets are utilized in their original, unmodified public formats.

\subsection{Node Generation Configuration}
In our experimental framework, we systematically controll the node generation process by setting a critical hyperparameter $k$. The number of node generation is constrained to $k = 3$, a carefully chosen value that strikes an optimal balance between comprehensive reasoning and computational efficiency.

\subsection{Node Selection Strategy}
Our node exploration methodology utilizes a structured three-tier categorization system, as illustrated in Table~\ref{tab:node-selection}. This hierarchical selection approach optimizes computational resources by focusing on the most promising reasoning pathways.

\begin{table}[h]
\centering
\begin{tabular}{|p{0.2\textwidth}|p{0.4\textwidth}|p{0.3\textwidth}|}
\hline
\textbf{Category} & \textbf{Description} & \textbf{Exploration Strategy} \\
\hline
``Sure'' & Nodes with high probability of leading to correct answers & Prioritized for immediate expansion \\
\hline
``Maybe'' & Nodes with uncertain but potential relevance & Conditionally explored based on context \\
\hline
``Impossible'' & Nodes demonstrably unlikely to contribute to solution & Immediately pruned from exploration path \\
\hline
\end{tabular}
\caption{Node selection categorization strategy.}
\label{tab:node-selection}
\end{table}

\section{Model Generalizability Experiments}
\label{model_generalizability}

To evaluate the generalizability of FGDIP, we conduct additional experiments on a range of language models beyond GPT-3.5-Turbo. The results demonstrate FGDIP's effectiveness across different model sizes and architectures.

\subsection{Experimental Setup}

We evaluate FGDIP on the HotpotQA dataset \citep{yang2018hotpotqa} using three additional language models: GPT-4o-mini, Gemini-1.5-flash, and GPT-4o. The experimental setup and evaluation metrics are consistent with those described in Appendix~\ref{Experimental_Configuration}.

\subsection{Results}
Table~\ref{tab:app_model_generalizability} presents the F1 scores achieved by different methods on the HotpotQA dataset across the evaluated models.
\begin{table}[!htbp]
\centering
\begin{tabular}{lccc}
\hline
\textbf{Method} & \textbf{GPT-4o-mini} & \textbf{Gemini-1.5-flash} & \textbf{GPT-4o} \\
\hline
Standard      & 21.69      & 22.36            & 45.50  \\
CoT           & 38.72      & 32.30           & 58.82  \\
UALA          & 42.10     & -            & 59.95  \\
FGDIP (Ours)   & \textbf{45.37}  & \textbf{45.62}            & \textbf{65.69}  \\
\hline
\end{tabular}
\caption{Comparison of F1 score performance of different methods on the HotpotQA dataset (\%). UALA's reliance on the logprobs parameter restricts its use to specific models like the GPT series. Our framework is applicable to other models like Claude and Gemini.}
\label{tab:app_model_generalizability}
\end{table}
The results demonstrate that FGDIP consistently outperforms baseline methods across all tested models.  On smaller models like GPT-4o-mini and Gemini-1.5-flash, FGDIP achieves F1 scores of 45.37\% and 45.62\%, respectively, showcasing its effectiveness even with limited computational resources. In addition, the performance improvement on Gemini-1.5-flash is particularly obvious, which shows the potential of FGDIP on small and medium models.
On the larger GPT-4o model, FGDIP achieves an F1 score of 65.69\%, a significant improvement over its performance on GPT-3.5-Turbo, indicating that FGDIP scales effectively with model size. Larger models are better able to leverage the dynamic and adaptive reasoning capabilities of FGDIP.

Furthermore, unlike the UALA \citep{han2024uncertaintyawarelanguageagent} method, which is limited to models providing access to log probabilities (e.g., the GPT series), FGDIP demonstrates broad applicability across different model architectures. While Table~\ref{tab:app_model_generalizability} focuses on GPT series models for direct comparison with UALA, FGDIP's design allows it to be used with other models, such as Claude and Gemini.

\section{Impact of Module Removal on FGDIP Model Accuracy}

\subsection{Without the Extractor Module}
\label{Extractor Module}
\begin{table}[!htbp]
\centering
\begin{tabular}{lcc}
\hline
\textbf{Difficulty} & \textbf{FGDIP} & \textbf{w/o Extractor} \\
\hline
Easy & 43.42\% & 30.26\%$_{\downarrow 13.16\%}$ \\
Medium & 45.07\% & 32.39\%$_{\downarrow 12.68\%}$ \\
Hard & 38.16\% & 19.74\%$_{\downarrow 18.42\%}$ \\
\hline
\end{tabular}
\caption{Impact of removing the Extractor module on FGDIP model accuracy across different difficulty levels. The term w/o Extractor refers to the FGDIP model without the Extractor module.}
\label{tab:rates1}
\end{table}

\subsection{Without the Step Evaluator Module}
\label{Step Evaluator Module}
\begin{table}[h]
\centering
\begin{tabular}{lcc}
\hline
\textbf{Difficulty} & \textbf{FGDIP} & \textbf{w/o Step Evaluator} \\
\hline
Easy & 49.23\% & 52.31\%$_{\uparrow 3.08\%}$ \\
Medium & 50.94\% & 56.60\%$_{\uparrow 5.66\%}$ \\
Hard & 38.98\% & 37.29\%$_{\downarrow 1.69\%}$ \\
\hline
\end{tabular}
\caption{Impact of removing the Step Evaluator module on FGDIP model accuracy across different difficulty levels. The term w/o Step Evaluator refers to the FGDIP model without the Step Evaluator module.}
\label{tab:rates2}
\end{table}

\subsection{Without the Answer Evaluator Module}
\label{Answer Evaluator Module}
\begin{table}[!htbp]
\centering
\begin{tabular}{lcc}
\hline
\textbf{Difficulty} & \textbf{FGDIP}&\textbf{w/o Answer Evaluator} \\
\hline
Easy & 50.00\% & 42.42\%$_{\downarrow 7.58\%}$ \\
Medium & 46.59\% & 37.50\%$_{\downarrow 9.09\%}$ \\
Hard & 36.59\% & 31.71\%$_{\downarrow 4.88\%}$ \\
\hline
\end{tabular}
\caption{Impact of removing the Answer Evaluator module on FGDIP model accuracy across different difficulty levels. The term w/o Answer Evaluator refers to the FGDIP model without the Answer Evaluator module.}
\label{tab:rates}
\end{table}

\clearpage
\section{Cost}

\subsection{Token Cost}
\label{token cost}

\begin{table}[!htbp]
\centering
\begin{tabular}{l>{\centering\arraybackslash}p{2cm}>{\centering\arraybackslash}p{2cm}>{\centering\arraybackslash}p{2cm}>{\centering\arraybackslash}p{2cm}c}
\hline
\multirow{2}{*}{\centering\textbf{Method}} & \multicolumn{4}{c}{\textbf{HotpotQA}} \\
\cline{2-5}
 & Easy& Medium&Hard & Average  \\
\hline
Standard \citep{brown2020language} &216 & 205& 205& 209\\
CoT \citep{wei2023chainofthought} & 590 & 578 & 577 & 582 \\
ReAct \citep{yao2022react}  & 16009 & 17392 & 16696 & 16699\\
Reflexion \citep{shinn2023reflexion}  & 31278 & 31754 & 31413 & 31422\\
FGDIP (Ours)  & 117110 & 126088 & 126963 & 123387 \\
\hline
\end{tabular}

\caption{Cost of tokens for each task to various baselines on HotpotQA dataset.}
\label{tab:token cost}
\end{table}

\subsection{Time Cost}
\label{Time cost}

\begin{table}[!hbp]
\centering
\begin{tabular}{lc}
\hline
\textbf{Method} & \textbf{Cost} \\
\hline
Standard & 0.98  \\
CoT & 1.55  \\
ReAct & 23.70  \\
Reflexion& 44.91\\
FGDIP(Ours)& 100.88\\
\hline
\end{tabular}
\caption{Time cost comparison of different methods on the HotpotQA dataset (in seconds).}
\label{tab:costs}
\end{table}

\section{Prompt}
\label{sec:appendix}
\subsection{Prompt to Extract Key Entities }

\begin{longtable}{p{0.9\textwidth}}
\hline
\noindent Extract one or more key entities based on a multi-hop problem so that the agent can try to solve the problem starting from a certain key entity.
Your task is to identify and extract well-defined entities in a complex information environment. When faced with a problem that contains multiple entities, you need to distinguish which entities are specific and unique, and which entities may refer to broad concepts or categories. Directive entities are usually individually identifiable names of people, places, specific events or unique objects, and they have clear definitions and boundaries in the context. In contrast, generic entities are more vague, such as common nouns or broad categories that can cover many specific instances. Your task is to focus on those clear entities, identify their role and importance in a given problem, and distinguish them from broader or vague concepts.\\
\noindent Here are some examples:\\
\noindent Question: Geoff LaTulippe is an American writer whose best-known work was directed by whom?\\
\noindent ["Geoff LaTulippe"]\\
\noindent Question: What is the population of the city which  Deryl Dedmon came from?\\
\noindent["Deryl Dedmon"]\\
\noindent Question: Who is older Danny Green or James Worthy?\\
\noindent ["Danny Green","James Worthy"]\\
\noindent Question: Alfie Allen played Theon Greyjoy on which show?\\
\noindent ["Alfie Allen","Theon Greyjoy"]\\
\noindent Question: The Bragg–Gray cavity theory was developed by Louis Harold Gray, William Lawrence Bragg, and a man that was knighted in what year?\\
\noindent ["Bragg–Gray cavity theory", "Louis Harold Gray", "William Lawrence Bragg"]\\
\noindent Question: Which battle took place during the American Civil War, Battle of Gettysburg or Battle of Tarawa?\\
\noindent ["American Civil War", "Battle of Gettysburg", "Battle of Tarawa"]\\
\hline
\end{longtable}

\captionof{table}{Prompt used for extracting key entities.}

\subsection{Prompt to Evaluate Step}
\begin{longtable}{p{0.9\textwidth}}
\hline
\noindent Consider a series of steps involving thought, action, and observation that have been undertaken. Assess the likelihood that continuing such steps over time would lead to the correct answer (sure/maybe/impossible), instead of evaluating the likelihood of getting the answer right away in the next step. \\
\noindent For every case, you must think firstly in Judgement, and then output your Reply, with the format: \\
\noindent Judgment: (your analysis here). \\
\noindent Reply: sure/maybe/impossible \\
\noindent Here are some examples: \\
\noindent Question: Author David Chanoff has collaborated with a U.S. Navy admiral who served as the ambassador to the United Kingdom under which President? \\
\noindent Thought 1: I need to search David Chanoff, find the U.S. Navy admiral he collaborated with, then find the President under whom the admiral served as the ambassador to the United Kingdom. \\
\noindent Action 1: Search[David Chanoff] \\
\noindent Observation 1: David Chanoff is a noted author of non-fiction work. His work has typically involved collaborations with the principal protagonist of the work concerned. His collaborators have included; Augustus A. White, Joycelyn Elders, Đoàn Văn Toai, William J. Crowe, Ariel Sharon, Kenneth Good and Felix Zandman. \\
\noindent Thought 2: David Chanoff collaborated with William J. Crowe. I need to search William J. Crowe and find the President under whom he served as the ambassador to the United Kingdom. \\
\noindent Action 2: Search[William J. Crowe] \\
\noindent Observation 2: William James Crowe Jr. (January 2, 1925 – October 18, 2007) was a United States Navy admiral and diplomat who served as the 11th chairman of the Joint Chiefs of Staff under Presidents Ronald Reagan and George H. W. Bush, and as the ambassador to the United Kingdom and Chair of the Intelligence Oversight Board under President Bill Clinton. Crowe was born in La Grange, Kentucky on January 2, 1925. \\
\noindent Judgment: The question requires linking the collaboration of David Chanoff with a U.S. Navy admiral who also served as the ambassador to the United Kingdom, and then identifying under which president this admiral served in that diplomatic role. Based on the search results, David Chanoff collaborated with William J. Crowe, who is a well-documented figure. William J. Crowe served as the ambassador to the United Kingdom under President Bill Clinton. The dataset provides a clear lineage of roles and responsibilities for Admiral Crowe, connecting his military and diplomatic careers to his collaboration with David Chanoff. Thus, the contex is comprehensive and direct enough to affirmatively answer the question. \\
\noindent Reply: sure \\ \\
\noindent Question: What nationality is the man who developed the dictator game? \\
\noindent Thought 1: I need to search the dictator game, find the man who developed it, then find his nationality. \\
\noindent Action 1: Search[dictator game] \\
\noindent Observation 1: The dictator game is a popular experimental instrument in social psychology and economics,[1] a derivative of the ultimatum game. The term game is a misnomer because it captures a decision by a single player: to send money to another or not.[2] Thus, the dictator has the most power and holds the preferred position in this “game.” Although the “dictator” has the most power and presents a take it or leave it offer, the game has mixed results based on different behavioral attributes.[3] The results – where most dictators choose to send money – evidence the role of fairness and norms in economic behavior, and undermine the assumption of narrow self-interest when given the opportunity to maximise one's own profits.[4]. The dictator game is a derivative of the ultimatum game, in which one player (the proposer) provides a one-time offer to the other (the responder). The responder can choose to either accept or reject the proposer's bid, but rejecting the bid would result in both players receiving a payoff of 0. In the dictator game, the first player, the dictator, determines how to split an endowment (such as a cash prize) between themselves and the second player[5] (the recipient). \\
\noindent Thought 2: The paragraph does not mention the man who developed the dictator game. I need to look up developed. \\
\noindent Action 2: Lookup[developed] \\
\noindent Observation 2: (Result 1 / 1) The initial game was developed by Daniel Kahneman in the 1980s and involved three parties, with one active and two passive participants. \\
\noindent Judgment: The steps started with researching the dictator game, leading to a description of the game but not the developer. The follow-up search on the development of the game revealed Daniel Kahneman as the developer. The next logical step would be to determine Kahneman's nationality. Kahneman is well-known in the field of psychology and economics, and finding his nationality from available biographical information is straightforward. \\
\noindent Reply: sure \\ \\
\noindent Question: The term snake eyes is used in a particular casino game that uses what game pieces? \\
\noindent Thought 1: I need to search snake eyes, find the casino game it is used in, then find what game pieces it uses. \\
\noindent Action 1: Search[snake eyes] \\
\noindent Observation 1: Snake eyes is a roll of two dice, with one pip on each dice. Snake Eyes may also refer to:. 1 Film. 2 Music. 3 Literature. \\
\noindent Thought 2: The paragraph does not mention the casino game that uses snake eyes. I need to look up casino game. \\
\noindent Action 2: Lookup[casino game] \\
\noindent Observation 2: No more results. \\
\noindent Thought 3: I need to search snake eyes casino game instead. \\
\noindent Action 3: Search[snake eyes casino game] \\
\noindent Observation 3: Could not find snake eyes casino game. Similar: ['Craps (redirect from Snake-eyes)', 'Harrah's Entertainment (redirect from Harrah's Casino)', 'Atlantic City, New Jersey (redirect from Casinos in Atlantic City)', 'List of Nintendo Entertainment System games (redirect from Casino Derby and Super Bingo)', 'Dice (redirect from Game dice)']. \\
\noindent Judgment: The initial steps to identify the casino game associated with the term "snake eyes" led to the discovery that it involves a roll of two dice. However, there was confusion about which specific game uses this term, prompting further searches. The action to search specifically for "snake eyes casino game" and encountering redirects suggests a focus on dice games. While the observation does not directly state it, "Craps" is the implied game associated with "snake eyes," a term used when rolling two ones with dice in this game. \\
\noindent Reply: maybe \\ \\
\noindent Question: The History of Ranching is a mural that was originally located at an American brewery that in 1999 began trasferring its production to who? \\
\noindent Thought 1: I need to search The History of Ranching mural, find the American brewery it was originally located at, then find who it began transferring its production to in 1999. \\
\noindent Action 1: Search[The History of Ranching] \\
\noindent Observation 1: Could not find The History of Ranching. Similar: ['Ranch', 'The History of Ranching (Winn)', 'The Ranch', '6666 Ranch', 'The Secret of Skinwalker Ranch']. \\
\noindent Thought 2: To find the American brewery, I can search The History of Ranching (Winn). \\
\noindent Action 2: Search[The History of Ranching (Winn)] \\
\noindent Observation 2: The History of Ranching is a mural that was originally located at the Pearl Brewery in San Antonio, Texas. The Pearl Brewing Company had always preferred to use Texas artists whenever possible. Long after other brewers had moved to using only photographs in their advertising campaigns, Pearl continued to use ads that featured sketches, drawings, or paintings of the concept or product. The more famous of Pearl's artistic advertisements were of Judge Roy Bean's Jersey Lilly and a collage called The Last Fight of Manolete. Both works saw wide distribution as large printed pictures framed and intended to hang in pubs and bars. \\
\noindent Judgment: The process began by searching for information about "The History of Ranching" mural, which initially did not yield specific results. After refining the search to include the name "Winn," information about the mural being located at the Pearl Brewery in San Antonio, Texas, was uncovered. The next natural step would be to search specifically for the transfer of production by Pearl Brewery in 1999. Given the clear identification of the brewery and the specific timeframe, finding out who they began transferring production to is feasible but might require detailed historical business information about the brewery. \\
\noindent Reply: maybe \\ \\
\noindent Question: The song Justice was used for a WWE event that took place on what day? \\
\noindent Thought 1: This is a what type of question, and the clue is song Justice, WWE event and day. So I need to search song Justice and WWE event, find the WWE event that used the song Justice, then find the day the event took place. \\
\noindent Action 1: Search[song Justice] \\
\noindent Observation 1: Could not find song Justice. Similar: ['Cross (Justice album)', 'Stress (Justice song)', 'Poetic Justice (song)', 'Justice for All (song)', '.And Justice for All (song)']. \\
\noindent Thought 2: I can search Cross (Justice album) since it is the most similar. \\
\noindent Action 2: Search[Cross (Justice album)] \\
\noindent Observation 2: Cross (stylized as †; titled Justice on digital platforms[1]) is the debut studio album by French electronic music duo Justice. It was first released digitally on 11 June 2007,[1] and later on 18 June through Ed Banger Records and Because Music in most countries and Vice Records in the United States. Recorded during 2005 and 2006 in Paris, Cross was composed as an "opera-disco" album. It features many samples and "microsamples" throughout, with about 400 albums being used as sampled material. These include samples from Prince, Britney Spears and Madonna.[2] The song "D.A.N.C.E." is a tribute to Michael Jackson. \\
\noindent Thought 3: Cross (Justice album) was released on 11 June 2007. I need to search WWE event next and find the event that used the song Justice. \\
\noindent Action 3: Search[WWE event] \\
\noindent Observation 3: Could not find WWE event. Similar: ['List of WWE pay-per-view and livestreaming supercards', 'WWE Main Event', 'Royal Rumble (redirect from Wwe royal rumble)', "Saturday Night's Main Event", 'WWE Clash at the Castle']. \\
\noindent Thought 4: I can search Royal Rumble since it is the most similar. \\
\noindent Action 4: Search[Royal Rumble] \\
\noindent Observation 4: The Royal Rumble is a professional wrestling event, produced annually in late January since 1988 by WWE, the world's largest professional wrestling promotion. It is named after the Royal Rumble match, a modified battle royal in which the participants enter at timed intervals instead of all beginning in the ring at the same time. After the inaugural 1988 event aired as a television special on the USA Network, the Royal Rumble has been broadcast via pay-per-view since the 1989 event and livestreaming since the 2015 event. It is one of WWE's five biggest events of the year, along with WrestleMania, SummerSlam, Survivor Series, and Money in the Bank, referred to as the "Big Five". The Royal Rumble match is generally held as the main event match of the annual event. \\
\noindent Judgment: The process to find the exact date of a WWE event that used the song "Justice" started with a search for the song itself, which did not directly yield relevant results. The link to the album "Cross" by Justice was established, but no connection to any specific WWE event was made. Further searches led to general information about WWE events, particularly the Royal Rumble, but did not provide specifics about the use of the song "Justice" at any WWE event. Without more precise or direct information linking the song to a specific event, determining the exact day remains highly uncertain. \\
\noindent Reply: impossible \\ \\
\noindent Question: When was the female character that on June 9, 2015 was promoted to a series regular first introduced on "Once Upon a Time"? \\
\noindent Thought 1: I need to search "Once Upon a Time" and find when the female character was first introduced. \\
\noindent Action 1: Search["Once Upon a Time"] \\
\noindent Observation 1: "Once upon a time" is a stock phrase used to introduce a narrative of past events. Once Upon a Time may also refer to:. 1 Film. 1.1 Television. 2 Literature. \\
\noindent Thought 2: The paragraph does not mention the female character. I need to look up "female character". \\
\noindent Action 2: Lookup[female character] \\
\noindent Observation 2: No more results. \\
\noindent Thought 3: I need to search "Once Upon a Time" and female character instead. \\
\noindent Action 3: Search["Once Upon a Time" female character] \\
\noindent Observation 3: Could not find "Once Upon a Time" female character. Similar: ['Jennifer Morrison (category Female models from Illinois)', 'Nurse Ratched (category Once Upon a Time (TV series) characters)', 'Emma Swan (category Once Upon a Time (TV series) characters)', 'Red Queen (Through the Looking-Glass) (category Female characters in literature)', 'Elsa (Frozen) (redirect from Elsa (Once Upon a Time))']. \\
\noindent Thought 4: I need to search "Once Upon a Time" and Emma Swan. \\
\noindent Action 4: Search["Once Upon a Time" Emma Swan] \\
\noindent Observation 4: Could not find "Once Upon a Time" Emma Swan. Similar: ['Emma Swan', 'Swan Song (Once Upon a Time)', 'The Dark Swan', 'Tallahassee (Once Upon a Time)', 'Once Upon a Time (season 5)']. \\
\noindent Thought 5: I need to search "Once Upon a Time" and Emma Swan (Once Upon a Time). \\
\noindent Action 5: Search["Once Upon a Time" Emma Swan (Once Upon a Time)] \\
\noindent Observation 5: Could not find "Once Upon a Time" Emma Swan (Once Upon a Time). Similar: ['Emma Swan', 'Swan Song (Once Upon a Time)', 'The Dark Swan', 'Tallahassee (Once Upon a Time)', 'Lost Girl (Once Upon a Time)']. \\
\noindent Judgment: The search process aimed at identifying when a female character who was promoted to a series regular on June 9, 2015, in "Once Upon a Time" was first introduced faced significant challenges. The initial search provided information about the TV series but did not specify the character. Subsequent searches for the female character using various associated terms did not yield direct information about her first appearance date. The efforts included several terms related to potential characters but none successfully pinpointed the exact episode or date of introduction of the character being queried. \\
\noindent Reply: impossible \\
\hline
\end{longtable}
\captionof{table}{Prompt used for step evaluation.}

\subsection{Prompt to Evaluate Answer}
\begin{longtable}{p{0.9\textwidth}}
\hline
\noindent You will be given QUESTION and Get\_answer. You need to check whether the type of Get\_answer provided is consistent with the question. You need to identify the possible answer to the QUESTION, and then check whether the Get\_answer is the possible ANSWER. \\
\noindent For every case, you must think firstly in THOUGHT, and then output your JUDGMENT, with the format: \\
\noindent THOUGHT: (your analysis here). \\
\noindent JUDGMENT: YES / NO \\
\noindent Please note that the answer must directly answer the question. If the answer is 'unknown' or other words that indicate uncertainty and do not actually answer the question, then your judgment should be 'NO'. \\
\noindent Here are some examples: \\
\noindent Question: What is the elevation range for the area that the eastern sector of the Colorado orogeny extends into? \\
\noindent Get\_answer: 1,800 to 7,000 ft \\
\noindent THOUGHT: The eastern sector of the Colorado orogeny would have a specific elevation range, usually expressed in feet or meters. The given ANSWER "1,800 to 7,000 ft" specifies a range in feet, which is a plausible elevation range for an orogenic area. Therefore, this seems like a possible answer to the QUESTION about the elevation range of the area affected by the Colorado orogeny. \\
\noindent JUDGMENT: YES \\ \\
\noindent Question: Which continent have a distribution of both the Ternstroemia and Cussonia genus of plants? \\
\noindent Get\_answer: Africa \\
\noindent THOUGHT: The Ternstroemia and Cussonia genera of plants would be distributed in specific geographical regions, which could include one or more continents. The ANSWER "Africa" is given as the continent where both these genera are found. If both the Ternstroemia and Cussonia genera have a known distribution in Africa, then this answer would be correct. \\
\noindent JUDGMENT: YES \\ \\
\noindent Question: Where is the head quarter of the Blue Bird Body Company? \\
\noindent Get\_answer: Fort Valley, Georgia \\
\noindent THOUGHT: The Blue Bird Body Company is known for manufacturing school buses and other vehicles, and its headquarters location is a specific factual piece of information. The given ANSWER is "Fort Valley, Georgia," which, based on my existing knowledge, is correct. The Blue Bird Body Company is indeed headquartered in Fort Valley, Georgia. \\
\noindent JUDGMENT: YES \\ \\
\noindent Question: In what country did Gildo Siopaes win a bronze medal for bobsledding in the 1964 Winter Olympics? \\
\noindent Get\_answer: Italy \\
\noindent THOUGHT: The question asks about the country where Gildo Siopaes won a bronze medal in bobsledding during the 1964 Winter Olympics. The ANSWER provided is "Italy." However, there is issue here. Athletes win medals for their country, not in a country, unless it's asking about the location of the event. The 1964 Winter Olympics were held in Innsbruck, Austria, not Italy. It seems the question might be mixing details about an athlete's nationality with the location of the event or has incorrect details. \\
\noindent JUDGMENT: NO \\ \\
\noindent Question: The 1978 NBA World Championship Series had as MVP which Hall of Fame class member of 1988? \\
\noindent Get\_answer: yes \\
\noindent THOUGHT: The question is asking for the name of the person who was awarded MVP of the 1978 NBA World Championship Series and who was later inducted into the Hall of Fame in 1988. The correct answer should be the name of an individual, specifically a basketball player. The given ANSWER "yes" does not fulfill the requirement of naming the individual; it's merely an affirmation and does not provide the necessary information. \\
\noindent JUDGMENT: NO \\ \\
\noindent Question: Which Nobel laureate in Literature, known for exploring themes of existentialism, wrote a novel set during a plague in Algeria? \\
\noindent Get\_answer: Albert Einstein \\
\noindent THOUGHT: The question asks for a Nobel laureate in Literature who wrote about existential themes and set a novel during a plague in Algeria. The correct answer is Albert Camus, specifically for his novel "The Plague." Albert Einstein, on the other hand, was a physicist known for his contributions to the theory of relativity and quantum mechanics, and he was awarded the Nobel Prize in Physics, not Literature. Additionally, Einstein did not author novels. Thus, the ANSWER provided fails to correspond correctly to the question regarding a literary work and author. \\
\noindent JUDGMENT: NO \\
\hline
\end{longtable}
\captionof{table}{Prompt used for answer evaluation.}

\subsection{Prompt to Generate Nodes}
\begin{longtable}{p{0.9\textwidth}}
\hline
\noindent Focus on the following erroneous results and thinking that have been obtained to avoid solving the problem as before.Use them to improve your strategy of correctly answering the given question.

Results: {error results}, Thought: {thought}

...

\noindent  Based on other nodes on the same level and recent observations, you extract different key information, generate different thought, and thus generate different action.

Existing nodes:

Thought: {thought}\\ Action: {action}\\ Obs: {observation}\\
\hline
\end{longtable}
\captionof{table}{Prompt used for nodes generation.}

\end{document}